# Advanced Machine Learning and Deep Learning Techniques for Enhanced Cattle Identification and Detection: A Comprehensive Review

**Fayazunnesa Chowdhury[1], Syed Md. Galib[1,*], Md Nasim Adnan[1], Md. Moradul Siddique[1,2], Md Robiul Karim[3] and K M Tanvir Anjum[4]**

[1]Jashore University of Science and Technology, Jashore, Bangladesh
fayazunnesa.cse0405.c@diu.edu.bd; galib.cse@just.edu.bd; nasim.adnan@just.edu.bd
[2]University of Information Technology & Sciences (UITS), Dhaka, Bangladesh.
moradul_siddique@uits.edu.bd
[3]Gazipur Agricultural University, Gazipur, Bangladesh
vet_robiul@gau.edu.bd
[4]Shanto Mariam University of Creative Technology, Dhaka, Bangladesh
tanvirnp20@gmail.com
*Correspondence: galib.cse@just.edu.bd

**Abstract: The need for effective cattle identification technology is now more acutely felt than ever in maintaining biosecurity, food safety, and supply chain efficacy in livestock management. This paper presents a systematic review of recent research in cattle identification using machine learning and deep learning techniques. The present systematic review measures the effectiveness of traditional and modern cattle identification techniques using studies from major academic databases, where articles were subjected to full-text review. Among these techniques, classical Machine Learning Techniques such as K-Nearest Neighbors and Support Vector Machines have demonstrated good results in cattle identification; however, Deep Learning Techniques, such as Convolutional Neural Networks, Residual Networks, and You Only Look Once, are better in cognition, detection, and identification tasks. Feature extraction relies on common techniques like Local Binary Pattern (LBP), Speeded-Up Robust Features (SURF), and Scale-Invariant Feature Transform (SIFT), while key features commonly used in these studies include muzzle prints and coat patterns. The review highlights key hurdles involving cattle identification, such as the limited number of publicly accessible datasets, issues with data quality susceptible to environmental changes and animal mobility, and high demand for real-time processing ability. The paper aims to inform researchers, policymakers, and stakeholders about implementing scalable, humane, and effective cattle identification systems to achieve sustainable livestock management.**

**Keywords:** *Cattle Detection; Cattle Identification; Deep Learning; Livestock Management; Machine Learning*

---

## 1. Introduction

Cattle identification is a vital component of contemporary livestock management. It lays the groundwork for improved health monitoring, traceability, and breeding program optimization. The increased emphasis on sustainable agricultural techniques and the need to control livestock diseases have heightened the necessity for accurate identification. Growing demand for food safety also plays a role. Several identification strategies, ranging from traditional methods to modern technologies, have been devised in response [2]. These systems are essential for maintaining individual animal records and managing herds effectively. They help ensure livestock farming is both productive and sustainable [3]. Moreover, they play a critical role in the global food supply chain by enhancing the traceability and quality of meat and dairy products [4]. “Cattle identification” encompasses both 'cattle detection' and 'cattle recognition' as defined by [5]. Identification systems have progressed from manual to automatic methods using image processing. Conventional techniques, such as ear tagging [6], ear notching [7], and electronic devices [8], have all been used for individual cattle identification in farming.

Traditional cattle identification techniques include ear tags, branding, and tattooing. These have been used for many years and are relatively simple and cost-effective. However, they have significant drawbacks. These include susceptibility to loss or damage, potential harm to animal welfare, and tattoos eventually fading [9]. Modern techniques like RFID, facial recognition, and Machine Learning (ML) offer more reliable and ethical alternatives. Recent developments in image processing, ML, and DL have enabled advanced identification systems [10]. Image-based systems have gained attention for providing precise, non-intrusive, and automated cattle identification [11].

Visual trait-based techniques have long been used to identify and categorize cattle. They rely on distinct characteristics to recognize breeds or individual animals. In recent years, ML and DL technologies have been widely adopted for automatic identification using visual features [3, 12-15]. ML and DL, which are subfields of AI, address complex challenges for automated decision-making. ML systems learn from problem-specific training data and automate analytical technique building [16]. ML is usually divided into two main branches: supervised and unsupervised learning. Supervised learning uses labeled datasets, while unsupervised learning analyzes and groups unlabeled data. Unsupervised ML can detect patterns without human intervention [17]. DL is a machine learning concept based on Artificial Neural Networks [16]. Since 2012, deep learning has outperformed traditional ML in processing unstructured data such as images, video, speech, and text [18-20]. Its superior performance is due to its ability to automatically learn hierarchical features from raw pixel data, removing the need for manual feature engineering. However, traditional ML methods are still valuable. They are used with smaller datasets, require interpretability, or are applied where deep learning's computational needs are too high. The basic process in developing ML and DL systems includes a few main steps. First, a prepared dataset is used to train a model. Second, the model is validated using a separate validation dataset. Finally, the trained model is tested for performance on a test dataset. The data for ML includes features and labels. Features can be extracted using various techniques. DL can automatically extract high-level features and learn from them. Even though the development process seems basic, choosing the right technique, parameters, and feature set can be challenging for high prediction accuracy [17].

Various ML techniques, like K-Nearest Neighbors (KNN) and Support Vector Machines (SVM), have been used for cattle identification. These methods have had varying levels of success. They classify and identify animals using attributes from photos [21]. However, their performance often depends on the availability and quality of labeled data [22]. Despite these challenges, ML-based techniques help improve livestock identification procedures. Ongoing improvements in data collection and feature extraction are expected to make these systems even more reliable and accurate. The arrival of DL has further changed cattle identification methods. It has enabled the development of more robust and accurate techniques [2]. Widely used DL techniques include Convolutional Neural Networks (CNN), Faster R-CNN, Inception Networks, YOLO, and ResNet [23]. These excel at learning hierarchical features from raw image data, reducing the need for manual identification methods [18]. DL also handles large datasets and image variations well, making it suitable for practical applications [24]. These technological advances have streamlined cattle identification and supported better herd management, disease prevention, and traceability. As agriculture embraces innovation, robust cattle identification systems will be crucial in meeting the demands of modern livestock management.

### 1.1. Differentiation from Existing Reviews

Although some of the reviews have explored the computer vision in livestock management [4, 33-34], our work presents a unique novelty with five major contributions that fill the gaps in the literature:

**1.1.1. Feature-Centric Taxonomy**

Although some reviews have explored the role of computer vision in livestock management [4, 33-34], our work presents a unique novelty with five main contributions that fill existing gaps in the literature. These contributions are outlined as follows:

We classified our results by biometric modality: Muzzle, Coat, Facial, and Multi-Mode RGB-D. This contrasts with algorithm-focused reviews (Hossain *et al.* [33], Mahmud *et al.* [4]). Our approach helps practitioners select identification techniques based on operational constraints. For example, use muzzle recognition in managed settings and coat patterns for free-ranging herds. Our taxonomy measures feature-specific trade-offs. Muzzle prints show 93-100% F1 but are sensitive to occlusion. Coat patterns achieve 97-99.5% mAP with scalability, but they have 15-20% lighting variation [37, 47-48].

**1.1.2. Temporal Evolution Analysis**

We measured the 25-year adoption curve (2000–2025). The data shows a shift from traditional methods (90% to 40% adoption) to image-based methods (0% to 60% adoption). Current reviews did not cover this shift. Our analysis highlighted key turning points. Before 2018, single-feature muzzle-based approaches accounted for about 70%. Before 2020, multi-modal fusion exceeded 50%. Both were limited by data and computational resources. These trends guide research investments and policies on livestock identification infrastructure.

**1.1.3. Hardware-Performance Integration**

Our GPU-to-Edge deployment comparison addresses a gap in prior deployment reviews. Mahmud *et al.* [4] discussed features of the proposed algorithm (ResNet) but did not address computation feasibility for resource-constrained farms. We show that high-accuracy models (98-99% F1) often require faster hardware, such as the GPU Tesla P100 or GTX 2080. In contrast, edge devices like Jetson Nano or Raspberry Pi 4 deliver 92-96% accuracy with 10-50ms latency. This trade-off is critical for real-world deployment [48, 50, 55].

**1.1.4. Multi-Dimensional Trade-Off Analysis**

This review has combined contradictions in computer vision approaches. For example, YOLO achieves a 14.6ms inference time, while Mask R-CNN provides 96.1% segmentation accuracy but is 2-3x slower. We also quantified gaps in model accuracy. Multi-breed generalization causes a 20-30% mAP drop when models trained on Holstein are tested on Angus or indigenous breeds [132-135]. Descriptive reviews have not addressed these issues. Our analysis enables informed decisions. YOLO-class detectors are prioritized for real-time biosecurity. Mask R-CNN's computational cost is justified by its accuracy in precision breeding programs.

**1.1.5. Socio-Technical Framework**

We looked beyond technical reviews. We included ethical, economic, and legal considerations (Section 6) to highlight barriers to real-life deployment. We discussed cultural sensitivity, such as Qurbani needs of physical integrity, and economic feasibility, like a 15-20% reduction in large farms and ROI issues for smallholders. We addressed legal issues (GDPR-equivalent data privacy) and societal impacts, such as the digital divide affecting 40% of rural Bangladesh [136, 152]. Our holistic view bridges field adoption and laboratory performance.

## 2. Background Study on Cattle Identification

Cattle identification is a key practice for farm animals. It allows farmers and industry professionals to monitor individuals and make informed decisions about health, breeding, and resource allocation. As agriculture evolves, identification improves operational efficiency and supports animal welfare and food safety with traceability. Intelligent technologies have become vital for performance assessment and animal identification. These innovations help enhance productivity, animal health, and efficiency [26]. Recognizing each animal and keeping accurate records enable farms to prevent outbreaks, conserve resources, and boost productivity. This is important as the focus on sustainable, safe food production grows.

### 2.1. Traditional Techniques of Cattle Identification

Traditional cattle identification techniques have been used for a long time for Cattle within some herds. Examples of existing traditional techniques of animal identification are ear-tagging, ear notching, and branding, which may be effective but can risk the animal and have scaling challenges.

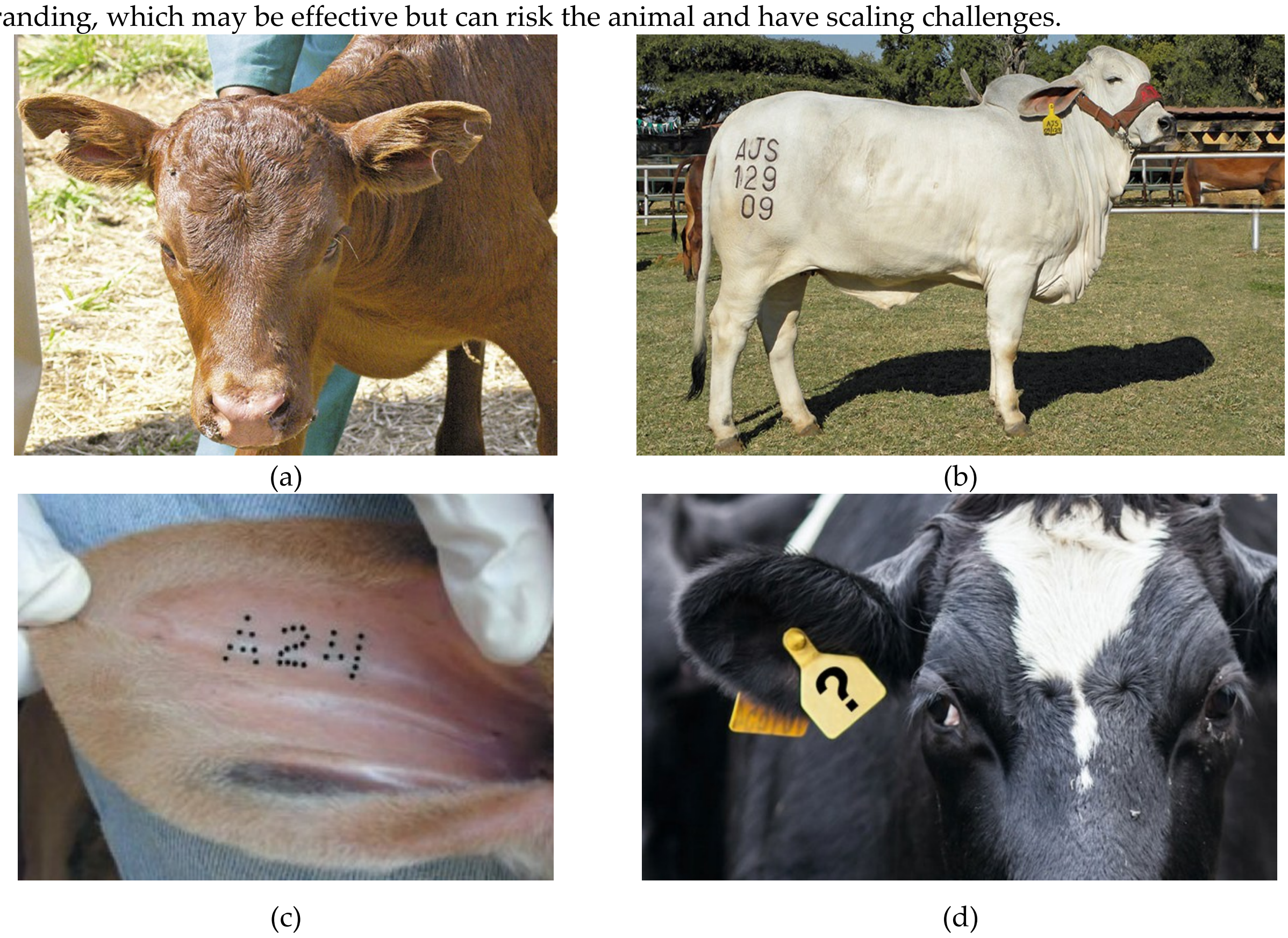

**Figure 1.** (a) Ear Notching [27], (b) Cattle Branding [28], (c) Tattooing of Cattle [29], (d) Ear Tags of Cattle [30]

These low-cost, simple techniques were invented to meet the farmer's needs in identifying, breeding, and freedom validation before modern technologies took over. Identification is done through physical markers, external tags, or permanent modification to parts of the animal's body. The most common manual techniques include:

#### 2.1.1. Ear Notching

Ear notching is a traditional technique used for herd identification. In this method, cattle are identified based on their birth order within available breeding. It provides permanent identification if done correctly [25]. The practice involves making distinctive patterns or notches along the edges of an ear to permanently mark it. Each notch corresponds to a code, number, or identification tag. This might indicate the birth year or herd number. Over the last few decades, ear notching has gained popularity in livestock farming, mainly due to its simplicity. Although it is one of the simplest and most economical identification techniques, it has several disadvantages. The process is painful and invasive, causing stress and discomfort to the animal. There is a risk of infection if the operation is not performed hygienically. The visual representation is insufficient for modern data-driven livestock management. The procedure is inefficient because the notch patterns must be created and recorded with care, especially in large herds. Over time, notches can become distorted or imprecise, leading to identification errors. In addition, the permanent damage to the ear may reduce the animal's value for shows or breeding. This makes ear notching less advantageous in modern livestock breeding.

In the context of Qurbani (sacrificial livestock), ear notching may not fulfil conventional or ethical requirements. Visible modifications to the animal's appearance can compromise physical integrity, which is a key criterion in Qurbani practices. Preserving the animal's welfare and unaltered appearance is essential in religious and cultural settings, indicating that ear notching is often not the preferred method.

Facial recognition systems using images do not require physical modification of cattle ears. They also perform better in identifying individuals. Recent deep learning methods with convolutional neural networks can identify cattle faces with over 95 percent accuracy, without physical interaction [40, 56]. This avoids the pain, infection risk, and aesthetic harm caused by ear notching. Multi-view facial recognition systems can recognize animals from various perspectives in free-stall barns. This helps overcome the scalability challenges of notching large herds (over 100 head). Non-invasive systems also preserve requirements for Qurbani because they do not damage the animal's physical integrity. This cultural need is not respected by ear notching.

#### 2.1.2. Branding

Branding has traditionally been used for livestock identification. It creates a permanent mark on an animal's skin to show ownership or identity. Hot branding (using a hot iron) or freeze branding (using extreme cold, such as liquid nitrogen or dry ice) are common methods. These create a visible mark on the hide. Branding has been used for centuries to establish ownership and avoid theft, especially with large, free-roaming herds. Hot iron branding is among the earliest cattle identification techniques. It dates back to the time of the ancient Egyptians [25].

Branding is reliable but has major downsides. The process is invasive and can cause pain, stress, burns, scarring, or infection. This raises serious animal welfare concerns. Branding is labor-intensive, requires skilled hands and proper tools, and takes time. Brands leave marks that can be longitudinal or steric, but they provide little information beyond the farmer of origin. Branding is not well-suited to data-rich herding or modern techniques. It can damage the animal's appearance, reduce market value, show potential, or suitability for Qurbani (sacrificial livestock). For Qurbani, branding does not align with traditional practices or expectations. Animals are expected to be in top physical and ethical condition. Over time, brands can become hard to read due to the environment or skin darkening, limiting their effectiveness.

Deep learning coat pattern recognition gives permanent, non-invasive identification. Unlike branding, it does not cause painful marking. State-of-the-art convolutional neural network (CNN) models reach 97-99.5% mean Average Precision (mAP) for coat pattern recognition [37, 48, 55]. These can use the unique pattern of each animal without burns, scarring, or stress. In contrast to brands that may fade or become unreadable due to hair growth or skin changes, image analysis is enhanced by larger datasets. Transfer learning improves model accuracy by reusing knowledge from similar tasks. Detection systems using the YOLO algorithm allow herd detection in pastures on the fly at 14.6-156 frames per second (fps) [50, 55]. Such performance is not achievable with traditional branding. These systems offer digital traceability along the supply chain and improve biosecurity and consumer confidence. They do this without reducing animal welfare, which is critical to Qurbani practices.

#### 2.1.3. Tattoos

Tattooing involves applying a permanent tattoo, usually a numeric code, inside the animal's ear. This is done with a tattooing instrument and ink. In many tribes, tattoos serve as ownership, family tree, or identification details. Once healed, the tattoo is a permanent identifier for the animal's life. Although tattooing is permanent and tamper-proof, it has disadvantages that restrict its use in modern livestock management. The procedure is invasive and can be uncomfortable or stressful for the animal. It carries a risk of infection unless done in sterile conditions. Tattoos can only be read up close. This is not easy for large herds, and it is not quick. The technique is labor-intensive and needs expertise and careful handling. Over time, tattoos can fade or become illegible as skin changes or ink disappears. Tattoos store very limited data, usually only simple numeric or alphanumeric sequences. They do not meet the needs of modern livestock data systems. Checking, reading, and recording tattoos during real-time cattle identification is exhausting and time-consuming [29]. For Qurbani (sacrificial livestock), tattooing is unsuitable. The invasive process and permanent skin changes conflict with ethical and traditional expectations for the animal's best condition. Visibility and appearance concerns lower its acceptability for Qurbani, as the animal's physical integrity and welfare are highly valued.

Deep learning-based muzzle print recognition gives permanence like tattooing but without surgery. It uses the unique ridge pattern of each animal's muzzle, similar to human fingerprints. State-of-the-art algorithms obtain 93-100 percent F1-score on muzzle recognition [47, 52-54]. This is higher than tattoo recognition, which decays as ink fades or skin changes. Automated muzzle systems can identify animals over a distance of 0.5-3 meters with regular RGB cameras or smartphones. Tattoos need close proximity and restraining the animal [48, 52]. Muzzle recognition allows identification during regular feeding or movement, causing less stress. Few-shot transfer learning methods in deep learning can recognize cattle using only 10-20 images per animal, which is much less labor and expertise than tattooing and is more accurate. In Qurbani settings, muzzle recognition does not harm physical integrity or animal welfare. Tattooing, in contrast, causes pain, risk of infection, and permanent changes to the skin, making it culturally inappropriate.

#### 2.1.4. Ear Tagging

Ear tagging is a common, conventional way to identify cattle. A tag, usually plastic or metal, is placed on the animal's ear. The tag contains an identification number, barcode, or sometimes data like breed, age, or vaccination history. This technique allows easy visual identification in a herd. It is convenient and low-cost, and avoids some problems of older techniques, such as animal distress and difficulty with human inspection [31]. A specialized applicator pierces the ear and secures the tag. Ear tagging is used in both small and large livestock operations because it is simple and inexpensive.

The disadvantages of ear tagging include susceptibility to loss or damage, as tags may fall out accidentally or degrade due to environmental conditions or animal behavior, resulting in misidentification and data loss. Moreover, tagging is a stressful and painful process for the animal, and if done improperly, can lead to infection of the piercing site and risk issues about the animal's welfare. Tags can also be tampered with, causing issues with identification validation in cases of ownership disputes or traceability needs. Ear tags have been found to be susceptible to damage, duplication, loss, unreadability, and fraud [29]. Moreover, ear tags alone do not perform well as a long-term identification technique [32]. Furthermore, ear tagging is not ideal for Qurbani (sacrificial livestock), as it does not align with the traditional practices and ethical considerations of ensuring the physical integrity and welfare of the animal during the sacrifice process.

Fusion techniques with facial, body and gait recognition have reached 96-99% F1-score [38, 40] in complete non-contact identification which removes all risks associated with ear tags: loss (tags falling out during grazing or handling), damage (environmental degradation of plastic/metal), swapping (fraudulent swapping of tags), and infection (piercing-site complications). Depth cameras used in RGB-D sensing systems (Kinect, Intel RealSense) can be utilized to identify the top-view without the need to touch the animal [39-42], which is particularly useful in the context of welfare-oriented operation and preparations of Qurbani, where the minimization of stress is the primary concern. Re-identification networks using deep learning can follow individual cattle across cameras and over time periods [50, 95], giving them a persistent digital identity, which can never be lost or erased, overcoming the inherent insecurity of external tag systems. In large-scale commercial use, IoT-based automated identification with cameras and farm management software makes it possible to have real-time traceability between birth and market of an animal with ease, better than ear tags due to their inherent information limits, without becoming unethical or culturally offensive.

#### 2.1.5. From Traditional to Modern: The ML/DL Paradigm Shift

The classical identification techniques have many technical constraints. These include animal welfare issues (pain, stress, injury), operational weaknesses (loss, damage, fading), cultural concerns (physical changes violating Qurbani prerequisites), and labor requirements (manual application and reading). These shortcomings directly led to the creation of image-based machine learning and deep learning algorithms examined in this paper. Contemporary solutions address these issues using non-invasive, automated, and ethically acceptable methods. Table 1 shows the limitations of conventional methods and the related ML/DL solutions.

**Table 1.** Traditional Method Limitations and Corresponding ML/DL Solutions

| Method | Primary Limitation | ML/DL Solution | Performance Achieved | Key Advantages | References |
|---|---|---|---|---|---|
| Ear Notching | Invasive, painful, permanent ear damage, infection risk | Facial recognition (CNN-based) | >95% accuracy, 0% physical contact | Non-invasive, no welfare concerns, scalable to large herds, Qurbani-compliant | [40, 56] |
| Branding | Severe pain (burns), scarring, slow application, brands fade over time | Coat pattern recognition (YOLO, ResNet) | 97-99.5% mAP, real-time detection | No pain/scarring, permanent digital record, real-time processing (14.6-156 fps), supply chain traceability | [37, 48, 55] |
| Tattoos | Pain, infection risk, fading, close-proximity reading required, labor-intensive | Muzzle print recognition (VGG, DenseNet, few-shot learning) | 93-100% F1-score, 0.5-3m reading distance | Permanent biometric (ridges unchanging), remote reading, no physical alteration, welfare-friendly | [47, 52-54] |
| Ear Tags | Loss (tags fall out), damage (environmental), tampering (fraud), infection (piercing), static data capacity | Multi-modal fusion (Face + Body + Gait, RGB-D depth sensing) | 96-99% F1-score, persistent digital identity | Cannot be lost/removed, tamper-proof, dynamic data integration, IoT-enabled real-time tracking | [38-42] |

### 2.2. Systematic Review Methodology

#### 2.2.1. Search Strategy and Selection Criteria

To ensure a comprehensive and reproducible review, we followed PRISMA guidelines. We conducted a systematic search across major academic databases. These include IEEE Xplore, ScienceDirect, Scopus, Web of Science, and SpringerLink. We used Boolean search terms to target the intersection of species and technology: ("cattle identification" OR "cattle detection" OR "cattle recognition") AND ("machine learning" OR "deep learning" OR "computer vision" OR "CNN" OR "neural network"). Our search was limited to English-language publications from 2000 to 2025.

Strict inclusion and exclusion criteria were used to screen the results. Included items were peer-reviewed articles, conference papers, and preprints using ML or DL techniques for cattle identification based on visual features. Papers were excluded if not in English, focused only on other livestock, were review articles, or focused on hardware implementations without strong algorithmic content.

#### 2.2.2. Study Selection and Data Extraction

The study selection process followed a multi-stage filtering approach. The initial search across all databases yielded a total of 487 papers. Following a primary screening of titles and abstracts to remove irrelevant entries and duplicates, 156 papers were retained for detailed examination. These candidates were subjected to a full-text review, resulting in 89 papers that met all inclusion criteria for the final analysis. The complete selection flow is illustrated in Figure 2.

For each selected study, we extracted key information such as method type (ML vs. DL), feature extraction techniques, dataset characteristics, accuracy metrics, publication year, and computational requirements. Additionally, we assessed each article's reliability using criteria including dataset size, experimental rigor, reproducibility, validation methods, and statistical significance reporting.

#### 2.2.3. Study Selection and Data Extraction

Furthermore, a comprehensive quality assessment was conducted for each article to evaluate the reliability and validity of findings. We employed a modified CASP (Critical Appraisal Skills Program) framework adapted for computer vision research in precision livestock farming. Each selected study was scored (0-2 points) across seven critical dimensions:

##### 2.2.3.1. Dataset Quality (0-2 points)

- **2 points:** Large-scale (>1,000 images), multi-breed diversity, publicly available with DOI/repository link, annotations verified by domain experts (veterinarians or animal scientists).
- **point:** Moderate-scale (100-1,000 images), single-breed or limited diversity, restricted availability (available upon request), standard annotation protocol.
- **0 points:** Small-scale (<100 images), breed undocumented, unavailable for validation, annotation methodology unclear.

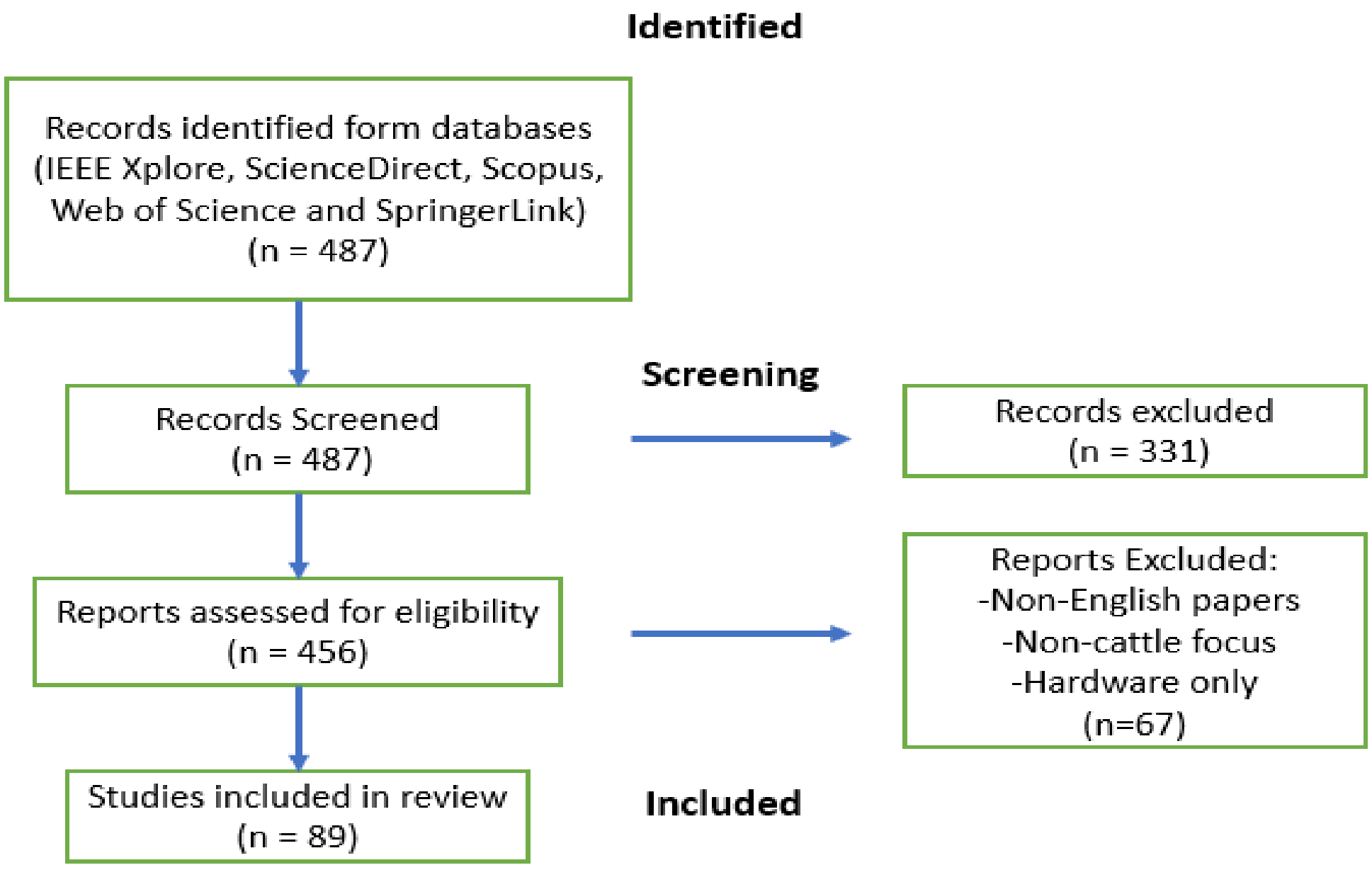

**Figure 2.** PRISMA flow diagram of the study selection process

#### 2.2.3.2. Experimental Rigor (0-2 points)

- **2 points:** Multiple independent train/test splits, k-fold cross-validation (k≥5), separate held-out test set never used in training, reported confidence intervals or standard deviations.
- **1 point:** Single train/test split (70/30, 80/20, or 60/40) clearly documented, validation set used.
- **0 points:** No validation methodology described, or training and testing on same data (methodological flaw).

#### 2.2.3.3. Reproducibility (0-2 points)

- **2 points:** Source code publicly available (GitHub/GitLab/Zenodo), all hyperparameters fully specified (learning rate, batch size, epochs, optimization algorithm), random seeds documented for reproducibility.
- **1 point:** Hyperparameters specified in detail but source code unavailable, sufficient implementation details for replication.
- **0 points:** Insufficient implementation details, hyperparameters missing or vaguely described ("default settings" without specification).

#### 2.2.3.4. Baseline Comparisons (0-2 points)

- **2 points:** Compared against ≥3 baseline methods (classical ML and contemporary DL), statistical significance testing performed (t-test, Wilcoxon, McNemar), ablation studies conducted.
- **1 point:** Compared against 1-2 baseline methods, no formal significance testing but clear performance differences.
- **0 points:** No baseline comparisons, standalone results only (cannot assess relative performance).

#### 2.2.3.5. Performance Metric Reporting (0-2 points)

- **2 points:** Multiple complementary metrics reported (Precision, Recall, F1-score, mAP, AUC-ROC), confusion matrix provided, per-class performance analyzed for multi-class scenarios.
- **1 point:** 2-3 metrics reported (e.g., Accuracy and Precision), sufficient for evaluation.
- **0 points:** Single metric only (accuracy alone, which can be misleading for imbalanced datasets) or incomplete metric reporting.

#### 2.2.3.6. Performance Metric Reporting (0-2 points)

- **2 points:** Tested on multiple farms with different infrastructure, diverse environmental conditions (indoor/outdoor, day/night, various weather), temporal validation across seasons or years.
- **1 point:** Single farm or laboratory setting, controlled environmental conditions, some variation documented.

- **0 points:** Simulated or synthetic data only, no real cattle images, or testing environment undocumented.

#### 2.2.3.7. Computational Transparency (0-2 points)

- **2 points:** Inference time reported (ms/image or FPS), hardware specifications detailed (GPU model, CPU, RAM), training time documented, energy consumption or carbon footprint reported.
- 1 point: Hardware mentioned (e.g., "GPU used") but limited performance metrics, training time provided.
- 0 points: No computational details, hardware-agnostic claims without validation.

### 2.2.4. Study Selection and Data Extraction

Studies scoring ≥8 out of 14 points were included in the final analysis. This threshold ensures moderate-to-high methodological quality while maintaining sufficient breadth for comprehensive review. Studies below this threshold were excluded during full-text review due to insufficient rigor. Quality Distribution of Selected Studies (n=89):

- **High quality (12-14 points):** 23 studies (26%) — Exemplary methodology, fully reproducible, multi-farm validation.
- **Moderate-high quality (10-11 points):** 31 studies (35%) — Strong methodology, minor reproducibility limitations.
- **Moderate quality (8-9 points): 35 studies (39%) —** Adequate methodology, some methodological gaps but valuable contributions.
- **Lower quality (<8 points, excluded):** 67 studies excluded during full-text review — Insufficient rigor, lack of validation, or incomplete reporting.

## 2.3. Solutions for these problems (Digital Techniques of Cattle Identification)

Traditional livestock identification methods face challenges, prompting a move towards advanced, non-invasive, and reliable technologies. Electronic identification has transformed livestock management by offering automated, precise, and efficient solutions that reduce manual labor and improve event tracing. However, price, data security, and technological dependency remain significant hurdles. Overall, these technologies represent a leap toward more efficient, humane, and data-driven livestock management.

### 2.3.1. Search Strategy and Selection Criteria

The advent of image-based ML (machine learning) and DL (deep learning) techniques paved the way for precise, effective, and highly scalable identification of individual cattle through the use of visual data. Through advanced techniques in ML and DL, these methods help to identify unique physical features of cattle, including coat patterns, facial structures, and body shapes. For the two primary applications of cattle discovery (locating cattle in images) and cattle distinguishing proof (identifying specific cattle), all the ML-based papers, in fact, address cattle distinguishing proof issues. However, both detection (finding cattle in images) and identification (telling individual cattle apart) problems were studied in the DL-based papers [33]. High-quality images or videos are processed to extract features and train models, allowing automated identification in real-time or batch-processing scenarios.

This technique has many advantages for the identification of cattle. It is non-invasive, minimizing stress and risk to the animals, and also obviates the need for physical identifiers like tags or collars. However, once trained with the right data, the techniques can achieve very high accuracy, even for mid to large-scale operations, while functioning effectively in a variety of environmental contexts. Incorporation with smart livestock management platforms can be done using image-based techniques, allowing the monitoring of health, behavior, and traceability. Also, these techniques are very flexible: they can be potentiated depending on breeds or management needs. Utilizing strategies in ML, image-based cattle identification is embedded as a game-changing solution for augmenting productivity, enhancing animal welfare, and streamlining operations within the livestock industry.

Image-based machine learning (ML) and deep learning (DL) techniques, such as convolutional neural networks (CNNs), have become the dominant approach in computer vision-based livestock identification tasks [34]. CNNs can automatically discern significant features by being trained on large numbers of cattle images and use these to form a permanent, non-invasive marker for each animal, eliminating the need for physical modifications or devices. This promotes better animal welfare and ethical and cultural compliance, especially in contexts like Qurbani, where maintaining the animal's physical integrity is essential.

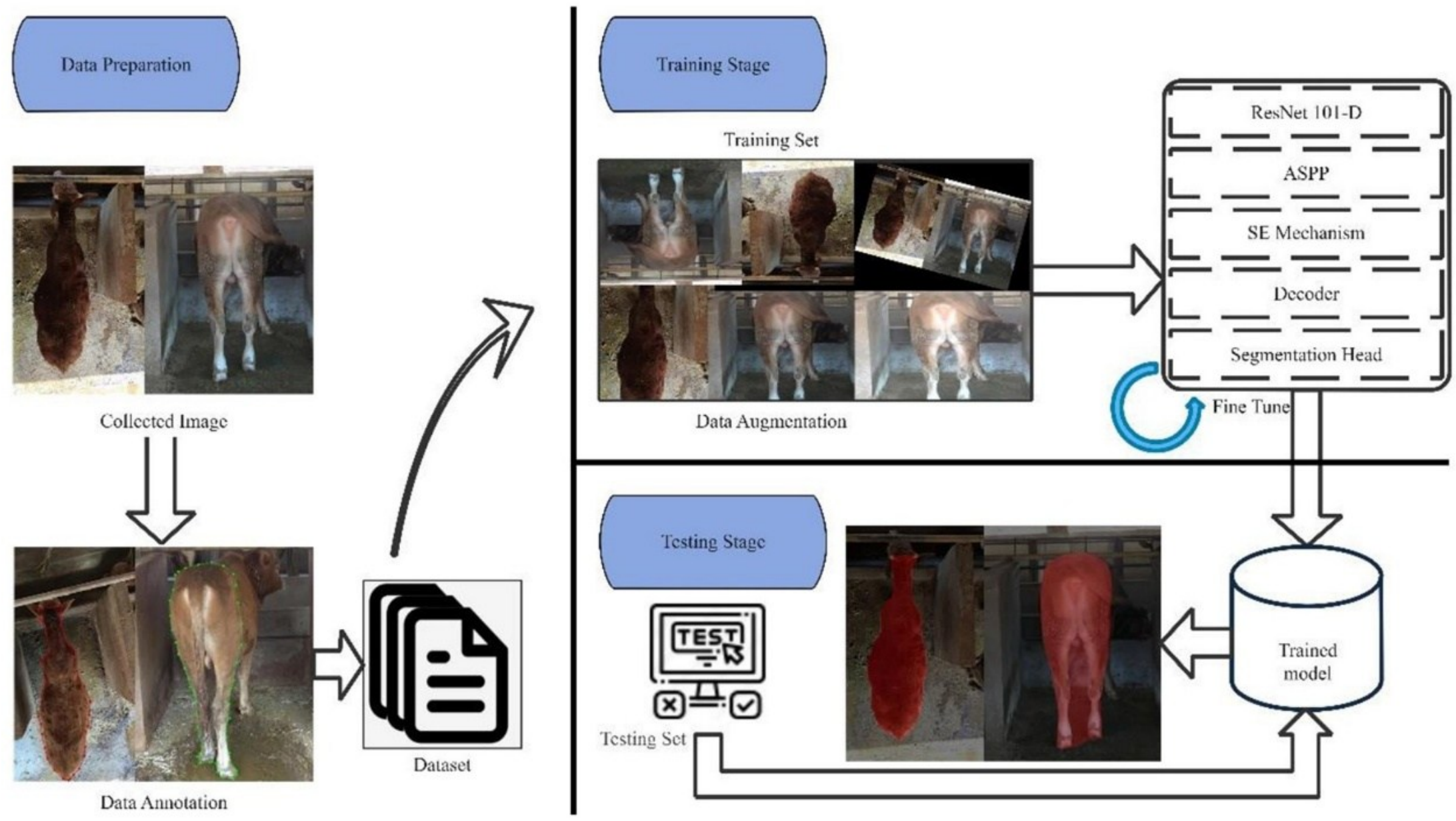

**Figure 3.** Image-based ML and DL Techniques Work procedure [35]

To Advanced deep learning (DL) techniques like ResNet (a residual neural network architecture), EfficientNet (a family of convolutional neural networks optimized for efficiency and accuracy), and YOLO (You Only Look Once, a real-time object detection algorithm) have shown significant promise in real-time object detection and recognition tasks. DL techniques like YOLOv5 (a version of the YOLO algorithm) have shown promise in real-time object recognition; their practical applicability may be constrained by their high processing requirements [36]. These techniques can process images from various angles, under different environmental conditions, and for a variety of breeds and herds. Integrating with AIoT (Artificial Intelligence of Things) devices like smart cameras or drones provides the ability for more scalable systems and real-time monitoring capabilities. AI-enabled systems can, for instance, automatically generate an individual record of each animal in free-ranging herds, which provides a distinct advantage in terms of savings in labor and operational efficiency.

Image-based ML and DL techniques can also incorporate multi-modal data—combining visual features with relevant metadata such as age, health status, or vaccination records in livestock management. Their flexibility allows application to small family farms as well as large commercial operations. By providing secure and tamper-proof identification, these methods improve traceability, verify ownership, and support regulatory compliance.

Head-based, Body-based, Muzzle-based, Coat-based, Shape-based, Tag-based, and Multiview patterns are some of the physical, biometric, and behavioral features used to identify cattle. Machine learning (ML) techniques, such as support vector machines and Random Forest, rely on features extracted through human effort (e.g., edges, texture), and deep learning (DL) techniques, such as CNN, automatically learn features from images through complex pipelines. Convolutional neural networks enable automatic feature learning through hierarchical representations, eliminating manual feature engineering required by traditional methods (e.g., ResNet) with techniques such as transfer learning and data augmentation that recognize features like hiding texture or gait. Real-time identification utilizing IoT Sensors combined with ML/DL improves traceability and monitoring.

A key distinction in image modalities is 2D RGB vs. RGB-D depth sensing: 2D excels in simplicity and low-cost deployment (e.g., standard cameras for coat patterns, >98% accuracy in daylight [37]), but struggles with pose/depth ambiguity in herds, dropping Recall by 10-20% [38]. RGB-D, leveraging Kinect/Intel RealSense for 3D point clouds, enhances identification by fusing depth with texture (e.g., gait-based re-ID at 96% F1 [39, 40]), mitigating occlusions and enabling top-view monitoring without handling—ideal for welfare-focused farms [41, 42]. However, RGB-D's higher compute (2x inference time) and sensitivity to sensor drift limit adoption to ~15% of studies, signaling a gap in hybrid 2D/3D frameworks for scalable, non-invasive systems [43-45].

### 3. Features of Cattle Identification

Cattle identification uses unique physical, biometric (measurable biological traits), and behavioral characteristics for rapid, large-scale recognition. This approach increases productivity and biosecurity. It also integrates Machine Learning (ML) and Deep Learning (DL) for better animal welfare and operations. For novelty, this review proposes a taxonomy organizing studies by input type or modality: Muzzle-based, Coat-based, Facial/Head-based, and Emerging Multi-Modal. Multi-Modal includes data from three-dimensional Red-Green-Blue-Depth (3D RGB-D) cameras. There is a shift from early single-feature focus (before 2018, about 70% muzzle-based due to its distinguishing patterns) to hybrid combinations (after 2020, with over 50% using multi-modal approaches). This shift improves performance across varying environments [17, 46]. This framework links feature choice to performance gaps. For example, muzzle features offer precision in controlled settings while coat-based features scale better in large herds.

Visual methods focus on key cattle body areas: muzzle, face, back, and trunk [39]. Early research used muzzle-based features such as patterns of ridges and grooves. Algorithms like Local Binary Patterns (LBP) and Scale-Invariant Feature Transform (SIFT) gave fingerprint-like uniqueness. Accuracy was above 98% in controlled laboratory data. However, performance dropped when features were blocked or dirty, for example, with occlusion or mud [47]. Coat-based features rely on color patterns or spots. They allow for scalable, non-contact identification using edge detection. Problems with lighting can lower F1-scores (a balance between precision and recall) by 15-20% compared to muzzle-based methods [37, 48]. Facial/Head-based features use landmarks such as eyes, nose, and ears. These methods can combine views from multiple angles. Deep Learning (DL) hybrid models, such as ResNet, improved recall (the rate of correct identifications) by 10% for animals in motion or changing poses [49, 50]. New multi-modal methods combine behavioral data (like gait or walking pattern) and 3D depth information. These help address the limits of 2D images, such as pose changes [38, 39]. In Figure 4, Head-based features refer to specific landmarks—eyes, nose, and ears—important for identification.

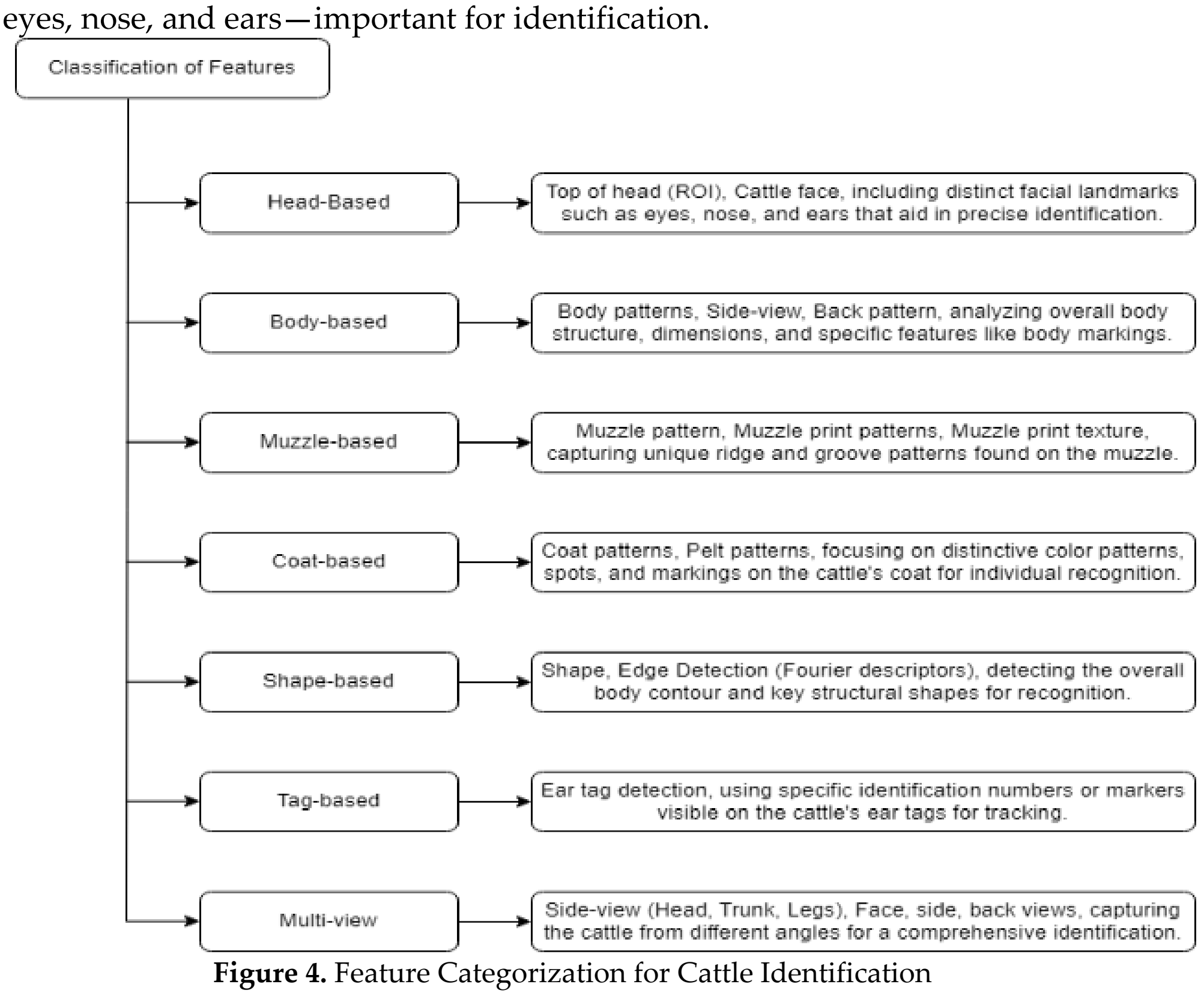

**Figure 4.** Feature Categorization for Cattle Identification

Body features help describe parts separately and interpret patterns such as side or back views, body structure, and visible body markings. Muzzle recognition uses the patterns of ridges and grooves found on a cow's nose. Coat recognition relies on unique color arrangements or spots on the animal's skin. Edge-detection algorithms find body outlines and shapes. These serve as shape features and help humans recognize cattle visually. Tag-based identification places physical markers, such as numbered ear tags, for easy identification. Multi-view analysis captures images from several angles—the head, trunk, and legs—for stronger identification. Previous reviews lacked: (a) organizing studies by biometric feature type (showing how each feature or modality performs), (b) quantifying how techniques or results change over 25 years, (c) providing information on hardware used and computational resource limits, (d) analyzing why identification accuracy may decrease across different cattle breeds, and (e) frameworks to describe social or regulatory barriers to adopting these systems. Our review addresses these gaps to guide researchers and practitioners. Table 2 compares our review with previous work.

**Table 2.** Traditional Method Limitations and Corresponding ML/DL Solutions

| Review Article | Year | Algorithm-Centric | Feature-Centric Taxonomy | Temporal Trends (Quantified) | Hardware-Deployment Context | Multi-Breed Generalization Analysis | Socio-Technical Framework | Quality Assessment (PRISMA) |
|---|---|---|---|---|---|---|---|---|
| Hossain *et al.* [33] | 2022 | ✓ Full | ✗ Not covered | ✗ Not covered | ✗ Not covered | ✗ Not covered | ✗ Not covered | Partial |
| Mahmud *et al.* [4] | 2021 | ✓ Full | ✗ Not covered | ✗ Not covered | ✗ Not covered | ✗ Not covered | ✗ Not covered | Partial |
| Meng *et al.* [34] | 2025 | ✓ Full | ✓ Partial (limited) | ✗ Not covered | ✗ Not covered | ✗ Not covered | ✗ Not covered | Partial |
| Our Review | 2026 | ✓ Full | ✓ Comprehensive | ✓ Quantified (2000-2025) | ✓ CPU/GPU/ Edge analysis | ✓ Quantified (20-30% mAP drop) | ✓ Compre-heensive | ✓ Full PRISMA compliance |

Table 3 benchmarked several key studies, revealing significance: Multi-modal yielded 5-15% mAP gains over 2D single-features in diverse datasets, underscoring the shift toward generalization [51].

**Table 3.** Master Comparative Framework: Quantitative Synthesis by Biometric Features

| Biometric Feature | Representative Studies | Dataset Size (Multi-Breed?) | Accuracy/F1/mAP | Inference Speed (ms) | Key Trade-Off |
|---|---|---|---|---|---|
| Muzzle | [47, 52-54] | 105-4,923 (Mostly single-breed Holstein) | 93-100% F1 | <50 (CPU) | High precision but occlusion-sensitive |
| Coat/Body | [37, 48, 55] | 377-1,965 (Limited multi-breed) | 97-99.5% mAP | 14.6-156 (GPU/Edge) | Scalable but lighting-variant |
| Facial/Head | [49, 50, 56] | ~505k frames (Multi-farm, partial multi-breed) | 84-98.4% mAP | 14.6-500 (Jetson Nano) | Dynamic but compute-heavy |
| Multi-Modal (3D RGB-D) | [38, 40, 57] | 301-1,448 segments (Emerging multi-breed) | 96-99% F1 | 50-200 (Edge) | Robust to pose but hardware-dependent |

Table 4 provides an overview of different cattle identification studies, specifying the datasets and feature-based techniques utilized. These techniques include head-based, body-based, muzzle-based, coat-based, shape-based, tag-based, and multiview methods. The studies apply a variety of datasets, such as images and videos. They utilize facial recognition, body patterns, muzzle prints, coat textures, shape detection, and multi-angle views. Modern neural techniques, including CNNs, YOLOv5, and DenseNet201, are used to improve identification accuracy.

**Table 4.** Overview of Cattle Identification Studies Based on Features and Techniques

| Ref. | Datasets | Head Based | Body-Based | Muzzle Based | Coat Based | Shape-Based | Tag Based | Multiview |
|---|---|---|---|---|---|---|---|---|
| [50] | Contains 281 video clips of 102 Holstein cows, focusing on facial recognition for identification | √ | | | | | | |
| [58] | Includes 7265 images of 17 Holstein dairy cows, focusing on the top of the head as a region of interest (ROI) | | √ | | | | | |
| [53] | Features 1000 images of 10 cow species, emphasizing body patterns for identification | | √ | | | | | |
| [54] | Provides 1433 images of 105 cows, focusing on side-view images for identification via deep learning | | √ | | | √ | | |
| [55] | Comprises data from Farm A, Farm B, and Farm C, focusing on back patterns of cattle | | √ | | | √ | | |
| [56] | Contains 4000 images of Holstein-Friesian cows, focusing on muzzle pattern identification | √ | | √ | | | | |
| [57] | Consists of 27,849 images of 50 cows, focusing on side-view identification using DenseNet201 | | √ | | | | | |
| [58] | Provides 660 images of various animals, including cattle, focusing on shape and edge detection for recognition | | √ | | | √ | | |
| [59] | Includes 958 images of 93 cows, focusing on the side-view head, trunk, and legs for identification | √ | √ | | | √ | | |
| [60] | Consists of 217 images of 31 cows, focusing on muzzle patterns for identification using multiple methods | √ | | √ | | | | |
| [61] | Contains 3,707 images of Holstein-Friesian cattle, focusing on coat patterns for identification | | √ | | √ | | | |
| [62] | Provides 4,736 instances from 46 Holstein-Friesian cattle individuals, focusing on coat pattern identification | | √ | | √ | | | |
| [63] | Contains 11,754 frames of 4 cows, focusing on pelt patterns for identification | | | | | √ | | |
| [52] | Includes 900 images of Wu Ling cattle, focusing on muzzle print texture identification | | | √ | | | | |
| [37] | Contains 4000 images of Holstein-Friesian cows, focusing on muzzle pattern identification using YOLOv5 | | | √ | | | | |
| [64] | Consists of 765 images of Kamphaeng Saen beef cattle, focusing on muzzle print patterns for identification | | | √ | | | | |
| [65] | Contains 460 images of cattle, focusing on muzzle pattern identification using SIFT and RANSAC | | | √ | | | | |
| [66] | Comprises video data focusing on ear tag recognition using head detection and CNN for digit recognition | √ | | | | | √ | |
| [67] | Includes 1087 images of Simmental cattle, focusing on face recognition using transfer learning | √ | | | | | | |
| [68] | Contains 27,849 images of Pantaneira cattle, focusing on face, side, and back views for multi-view identification | | √ | | | √ | | √ |

## 4. Performance Analysis

ML and DL were both key in cattle identification. Traditional ML, like SVM and Random Forests, excelled on structured datasets with manual feature engineering. DL methods, especially CNNs, automated feature extraction from unstructured images, delivering scalable, high-accuracy solutions. These approaches addressed various data complexities, from lab settings to dynamic farms, improving precision and efficiency. Over time, feature evolution shaped performance: early muzzle-focused ML (high Precision, low Recall for herds) gave way to DL-based coat and facial hybrid models, using multi-scale learning for occlusion resilience [37, 48].

### 4.1. ML Techniques in Cattle Identification

Classical ML approaches mainly targeted pattern recognition, image classification, and biometric analysis for cattle identification. These models emphasized non-intrusive, automated detection and outperformed manual verification in controlled settings. Trends showed strong use of texture features, like muzzle prints and body contours. Supervised classifiers processed handcrafted descriptors (e.g., LBP or SIFT) for robust individual matching, but performance often plateaued as dataset variability increased due to reliance on label quality.

**Table 5.** ML techniques used in cattle identification

| References | ML Techniques Used | Count |
|---|---|---|
| [41, 47, 45, 69-78] | Support Vector Machine (SVM) | 13 |
| [53, 79-83] | k-nearest neighbors (KNN) | 6 |
| [58, 66, 84, 85] | Artificial Neural Network (ANN) | 4 |
| [86, 87] | Decision Tree | 2 |
| [88, 52] | Linear Discriminant Analysis (LDA) | 2 |
| [89, 90] | BruteForce | 2 |
| [52, 78] | Naïve Bayes | 2 |
| [48] | Sparsity Residual Constraint | 1 |
| [35] | Random Forest | 1 |
| [35] | Logistic Regression | 1 |
| [44] | QDA | 1 |
| [79] | AdaBoost | 1 |

Table 5 highlights the thematic distribution of ML adoption. SVM stands out with over 12 applications and excels as a versatile classifier for high-dimensional biometric spaces, particularly when paired with local texture extractors that mitigate environmental noise in early herd monitoring studies. In contrast, proximity-based methods like KNN follow closely and offer practical simplicity for real-time classification of body or facial landmarks. Tree-based ensembles, such as Decision Trees and Random Forests, and dimensionality reducers like LDA, are moderately used for multi-view pattern clustering, which signals a shift toward hybrid feature fusion in mid-scale datasets. Less common techniques, such as AdaBoost or Naïve Bayes (each used less than four times), serve niche roles in anomaly detection or probabilistic modeling. Their limited adoption suggests opportunities for ensemble integration to enhance generalization across diverse breeds.

Analytically, ML techniques like SVM and KNN excelled with modest datasets (fewer than 1,000 images) and engineered features, delivering high precision (often above 95%) on benchmark metrics such as Precision, Recall, and F1-score in static biometric tasks. For example, muzzle-based studies showed SVM achieving near-perfect Recall, with a 99.7% F1 on tailhead datasets [47]. However, their dependence on manual feature crafting limited scalability for cattle-specific challenges, such as occlusion from herd clustering, variable farm lighting causing texture degradation, and breed variety leading to intra-class variability—for instance, differences in coat patterns between Holstein and indigenous breeds. These challenges increased false positives and dropped F1-scores below 80% in uncontrolled environments, with KNN's sensitivity to distance metrics also faltering under noise. In comparison to DL counterparts, ML's low computational footprint (for example, CPU-only inference in seconds) benefited resource-constrained smallholder farms but came with lower mAP (typically below 0.90 for multi-object detection). However, ML was insufficient for real-time, herd-scale applications, where DL's automated hierarchies yielded 10–20% gains in Recall and mAP.

### 4.2. ML Techniques in Cattle Identification

DL architectures, especially convolutional neural network (CNN) variants, have transformed cattle identification. They enable end-to-end learning of complex spatial hierarchies directly from raw imagery. This surpasses traditional machine learning (ML) in handling large amounts of unlabeled data for identifying multiple objects in herds. Key themes include moving from basic feedforward neural networks (which process data in one direction) to attention-driven models (which focus on important features throughout the image). The need to capture global contextual cues, such as coat variations hidden by overlapping animals (occlusion), drove this shift. Conventional local convolution operations often missed these global details, but attention models improved recall rates in real-world situations.

**Table 6.** DL techniques used in cattle identification [58]

| References | Identification Techniques | Count |
|---|---|---|
| [41, 55, 91-95] | Convolutional Neural Networks (CNN) | 7 |
| [43, 87, 96-99] | ResNet | 6 |
| [46, 87, 100-102] | Inception | 5 |
| [43, 48, 56, 88, 103] | VGG | 5 |
| [87, 104, 105] | DenseNet | 4 |
| [43, 106, 107] | AlexNet | 3 |
| [83, 108, 109] | Deep Belief Network (DBN) | 3 |
| [100, 110, 111] | SDAE | 3 |
| [112, 113] | Long-term recurrent convolutional network (LRCN) | 2 |
| [114, 115] | NasNet | 2 |
| [116, 117] | LeNet | 2 |
| [101] | PrimNet | 1 |
| [118] | Region-based Convolutional Neural Network (R-CNN) | 1 |
| [119] | DarkNet | 1 |
| [120] | Deep neural network (DNN) | 1 |

This frequency analysis (Table 6) revealed CNNs as the cornerstone (7 applications), favored for extracting local biometric features like muzzles. ResNet (6 uses) and Inception (5 uses) enhance this with residual skip connections and multi-scale modules to address vanishing gradients and improve breed differentiation. Moderate adoption of recurrent hybrids (e.g., LRCN) and autoencoder-based models (SDAE/DBN) suggests growing interest in temporal video modeling for re-identification. Less frequent use of foundational nets (e.g., AlexNet, LeNet) and specialized detectors (e.g., R-CNN, DarkNet) highlights a trend toward efficient backbones like DenseNet and VGG, indicating a shift to lightweight, interpretable designs for edge deployment under computational constraints.

**Table 7.** Comparison of Key Models in Cattle Identification

| Model | Technique Type | Use Case | Key Strengths | Importance |
|---|---|---|---|---|
| SVM | ML | Classifying based on features like face, body patterns | Handles high-dimensional data effectively | Widely used for classification tasks in structured datasets |
| KNN | ML | Identifying individual cattle via features | Simple and interpretable, effective for small datasets | Suitable for identifying animals based on unique traits |
| CNN | DL | Image-based identification (e.g., face, body) | Automatically extracts complex features, excels in large datasets | Best for handling unstructured data like images |
| ResNet | DL | Deep architectures for better performance in large datasets | Solves vanishing gradient issue, improves deep learning models | Enhances accuracy in deep architectures |
| YOLO | DL | Real-time detection and identification | Fast and efficient for real-time applications | Crucial for live detection in farm environments |

CNN, SVM, YOLO, and ResNet were the most used techniques for cattle identification (Table 7). These techniques showed promise because their efficiency varied according to the cattle identification tasks. This variation displayed considerable potential for specific applications depending on data complexity. Together, they enabled efficient solutions for a broad range of challenges, making them crucial in modern cattle identification systems. The importance of CNNs and ResNet lies in their capacity to handle large, unstructured image datasets. This minimizes the need for human intervention and optimizes precision in distinguishing cattle by unique visual features. SVM is powerful and straightforward for classification, and works best with controlled datasets and engineered features. YOLO, focused on speed, is perfect for real-time cattle tracking in situations where detection must be immediate. Combining these techniques can address identification, biosecurity, traceability, and herd management, resulting in sustainable livestock farming.

**Table 8.** Performance of ML and DL techniques used in cattle identification

| Reference | Year | Method | Feature | Dataset Size | Accuracy/mAP | Hardware Used |
|---|---|---|---|---|---|---|
| [121] | 2018 | SVM | Muzzle | 217 images | 100% | N/A |
| [37] | 2016 | SVM | Body | 377 images | 97% | Kinect 2 |
| [122] | 2018 | SVM | Mammary glands | 302 images | 60% | N/A |
| [47] | 2017 | QDA | Tailhead | 1,965 images | 99.7% | N/A |
| [123] | 2019 | SVM | Muzzle | 105 images | 93% | N/A |
| [48] | 2022 | VGG16_BN (transfer learning) | Muzzle | 4,923 images (268 cattle) | 98.7% | Tesla P100 GPU |
| [50] | 2022 | YOLOv5 + ResNet101 ArcFace | Face | 281 videos (~505,800 frames, 102 cows) | 84% | NVIDIA GTX 2080 / Jetson Nano |
| [124] | 2018 | CNN | Muzzle | >2,900 images | >98.9% | N/A |
| [53] | 2021 | Deep Belief Network (DBN) | Muzzle | >2,900 images | 98.9% | N/A |
| [52] | 2021 | Few-shot deep transfer learning | Muzzle | 2,900 images | 99.1% | N/A |
| [54] | 2023 | Custom CNN (VGGFace-based) | Muzzle | 4,923 images (268 cattle) | 100% (test) | Apple M1 Pro laptop |
| [55] | 2022 | YOLOv5 | Body | Not specified | mAP@0.5: 0.995 | N/A |
| [125] | 2024 | YOLOv8 + VGG16 + SVM | Back patterns | 1,448 images (multiple farms, ~1,263 cattle) | 96.34% (identification) | AXIS M3058-PLVE / P1448-LE cameras |
| [49] | 2024 | Various DL (e.g., CNNs) | Face | 4,243 images (OpenCows2020, 46 categories) | 98.42% | N/A |
| [56] | 2025 | BMLP preprocessing + DL recognition | Coat (black regions) | 301 video segments (153 train, 148 test) | ~7% improvement over baseline | N/A |

In In comparative terms (Table 8), deep learning (DL) models such as Convolutional Neural Networks (CNN) and Residual Networks (ResNet) outperformed traditional machine learning (ML) baselines like Support Vector Machines (SVM) and k-Nearest Neighbors (KNN) in cattle biometrics by automatically extracting invariant, hierarchical features from images. For example, these models can detect simple visual edges on cattle muzzles and progress to understanding complex breed-specific features—leading to higher F1-scores (such as 98-99% versus 90-95%) and mean Average Precision (mAP) scores over 0.95 on standardized datasets like OpenCows2020. This advantage comes from deep learning's specialized abilities: self-attention mechanisms in ResNet variants help minimize the effect of occlusion by considering context in partial views (resulting in a 15% increase in Recall for herds where cattle are partially blocked), multi-scale feature filters manage lighting changes (reducing performance drops caused by lighting from 20% in SVM models), and data augmentation increases breed identification robustness (with Precisions like 99.1% across mixed datasets). Specific network architectures also showed tailored benefits: Li *et al.* [48] tested 59 models and selected VGG16 with batch normalization (VGG16_BN) for muzzle print recognition, favoring its simpler architecture (16 layers vs. 50 in ResNet-50) to limit overfitting on moderately sized datasets (4,923 images) while batch normalization supported stable training for detailed textures and achieved 98.7% accuracy using 20% fewer parameters than ResNet, making it practical for farm-edge use.

By contrast, Dac *et al.* [50] incorporated ResNet101 as the ArcFace backbone within YOLOv5 for facial re-identification in videos (around 505,800 frames), using ResNet's residual links (which help maintain learning in deep 101-layer models) to identify cows in motion blur and herd overlaps, increasing mAP to 84%, though needing higher GPU resources (e.g., NVIDIA GTX 2080), which was justified by requirements for tracking cattle in large open yards. Shojaeipour *et al.* [52] adopted few-shot transfer learning with DenseNet, using its densely connected network structure for mixed-breed muzzle identification; this approach enhanced detection in small, diverse datasets (2,900 images), achieving 99.1% Precision by capturing subtle differences without needing exhaustive retraining. However, these benefits had trade-offs: deep learning models required far more computational power for training (10 to 100 times more, needing GPU hours compared to ML's minutes on a CPU), increased cost for deployment, and risked slower response (more than 500ms inference time on Jetson Nano vs. less than 50ms for KNN). Deep learning improved Recall (reducing false negatives due to cattle movement), but sometimes overfitted to dominant cattle breeds (such as Holstein), requiring hybrid techniques for balanced accuracy (mAP) across global breeds [53-56, 125].

The shift from basic CNNs to ResNet/YOLO hybrids showed a blend of feature modalities. Muzzle studies started end-to-end learning for fine textures [56, 124]. Research then evolved to coat-based YOLO for herd-scale detection [52, 55], where global attention bridged local biases. Contradictions arose in detector types: YOLO variants prioritized speed (14.6 ms inference, 97.4% mAP on body features [50]). Mask R-CNN offered higher accuracy for occluded counting (96.1% for sheep, but at 2-3x slower [53]), raising the debate of real-time biosecurity versus precise traceability. Debates also favored depth (thermal or RGB-D) for low-light (15% Recall boost [38]), although 2D was more efficient in clear conditions [31]. Hardware context influenced these trade-offs: cloud-deployed ResNets (e.g., Tesla P100 [55]) enabled over 99% F1 on large datasets but had a latency of over 500 ms. Edge variants on Jetson Nano or Raspberry Pi [54] sacrificed 5-10% mAP for under 50 ms, which was vital for offline farms [126].

#### 4.2.1. Emerging Architectures: Vision Transformers

Vision Transformers (ViT) are emerging in computer vision, especially for cattle identification. Unlike conventional CNNs, which use hierarchical feature maps, ViTs use self-attention to capture global image dependencies. This makes them highly effective for fine-grained tasks like cattle identification [127]. A recent study [128] introduced Multi-Head Attention Feature Fusion (MHAFF), which combines CNN and transformer strengths for cattle identification. This method replaces traditional addition and concatenation fusion with multi-head attention, enabling context-aware fusion. It achieved 99.88% and 99.52% accuracy on two cattle datasets. Another study [129] used Vision Transformers with transfer learning and Bi-former mechanisms for enhanced cattle face recognition. The Bi-Level Routing Attention enables the model to focus on different feature levels and improves attribute differentiation.

Further extending the application of ViTs, Vision Transformers were embedded in re-identification (ReID) networks like DeepSORT to enhance feature matching and tracking accuracy for autonomous livestock farming, addressing challenges of deformable body postures and irregular movements in complex farming environments [130]. While CNNs showed notable success in cattle identification using muzzle images with strong results, more recently, transformers like ViT and Swin Transformer have gained attention, demonstrating encouraging outcomes. Specifically, CNNs efficiently capture local features but struggle with long-range relationships, whereas transformer models handle global dependencies through self-attention but lack inherent inductive biases such as locality and shift invariance [131].

### 4.3. Challenges in ML and DL Adoption for Cattle Identification

Despite promising results, deploying deep learning (DL) and machine learning (ML) models faces domain-specific hurdles beyond familiar challenges. A key problem is limited data, which often means that models fail to generalize across breeds. For example, models trained mainly on Holstein cows (86% of US cattle data [132]) show a mean average precision (mAP) drop of 20–30% when tested on Angus or indigenous breeds. This mostly results from differences in coat and body structure. This challenges claims of 'universal' model backbones and points to gaps in reference data for crossbred cattle [133–135]. Environmental issues like occlusion (when animals or objects block each other) and mobility (animal movement) raise false negatives. For herd monitoring, recall (the proportion of true positives found) can drop 15–20% [136]. Night-time and low-light situations are not well studied. RGB-D (Red, Green, Blue plus Depth) sensors outperform standard 2D imaging by 10–15% because depth data better resists environmental variation [38, 39, 43]. Still, fewer than 10% of studies use RGB-D, often choosing the more expensive thermal imaging option instead [137].

Contradictions in model architectures fuel ongoing debate. For example, You Only Look Once (YOLO) is fast on edge devices (156 frames per second, fps [123]), but less accurate than Mask R-CNN in dense crowds, such as 80% versus 96% accuracy for cattle counting [47, 53]. These differences raise questions about YOLO's suitability for real-time use under variable lighting. Besides model selection challenges, deployment gaps add to these concerns. Cloud-based systems work well for training—such as using Amazon Web Services (AWS) for multi-modal fusion [48]—but are less used on edge devices. About 20% of academic papers tested deployment on barn-appropriate hardware like Raspberry Pi or NVIDIA Jetson [49, 124]. This can cause response delays of 50–100 milliseconds when offline, a critical flaw for remote farms [55, 56]. Annotation inconsistencies (expert labeling versus automated algorithms) bias results toward single-breed datasets. Combining human and AI annotation in hybrid pipelines is needed for fair scaling. Addressing these architectural, deployment, and annotation challenges is crucial for reliable, scalable livestock monitoring.

## 5. Trend Analysis

The performance of cattle identification systems employing machine learning or deep learning algorithms is typically assessed using quantitative metrics such as accuracy, precision, recall, F1 score, and computational efficiency. Figure 5 shows a trend analysis of cattle identification methods from 2000 to 2025. While the previous paragraph addresses performance metrics, it is helpful to understand how these methods have shifted over time. Traditional methods are steadily declining, dropping from 90% in 2000 to an estimated 40% by 2025 [29, 114]. Modern methods are rising. RFID (radio frequency identification) has grown from 10% in 2000 to 70% in 2025, becoming dominant [3, 115].

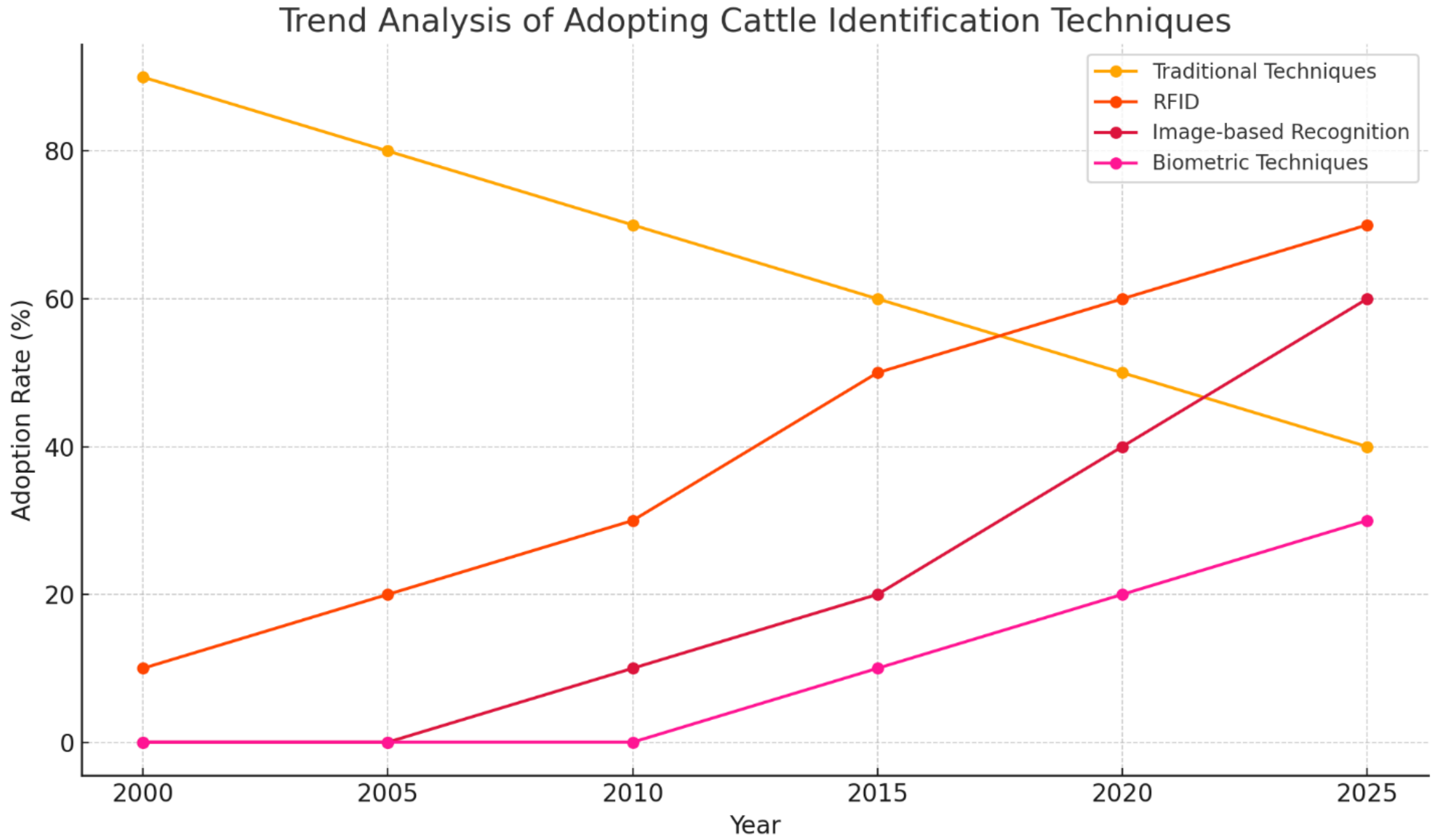

**Figure 5.** Trend analysis of cattle identification techniques from 2000 to 2025.

Image recognition, starting at 0%, is projected to reach 60% by 2025, driven by machine and deep learning advances [14]. Biometric methods (such as facial recognition and pattern-based ID) are emerging, increasing to 30% by 2025 [2, 5]. These changes show a shift to technology-driven identification. This trend suggests livestock management is becoming more precise, automated, and scalable, enabling improved herd monitoring and resource allocation [18].

### 5.1. Dataset Size Impact on Performance Comparability

Alongside these trends, one of the major challenges in cattle identification research is the wide variation in dataset sizes. Studies include datasets from 105 images [123] to 505,800 video frames [50], a nearly 5,000-fold difference. Comparing performance without considering the dataset context is misleading. Accuracy, F1-score, and mAP depend on the dataset scale. Small, controlled datasets often yield higher performance due to less variety. Large, real-world datasets feature more environmental diversity, temporal changes, and breed variation, which can be harder for models. For fair comparison, we suggest categorizing studies into four levels of dataset complexity by size, diversity, and realism:

#### 5.1.1. Tier 1: Small-Scale Datasets (< 500 images, n=15 studies)

- **Characteristics:** Laboratory-controlled environments, single-breed populations (typically Holstein-Friesian), limited environmental variance (consistent lighting, static camera angles, adult cattle only), short temporal collection periods (<1 month).
- **Typical ML Performance:** 93-100% (e.g., [121] achieves 100% on 217 muzzle images using SVM + LBP).
- **Typical DL Performance:** 85-95% (risk of overfitting due to insufficient training data for deep networks).
- **Best-Suited Methods:** Classical ML methods using handcrafted features (SVM, KNN, Random Forest) that leverage domain expertise.
- **Caveat for Interpretation:** High accuracy often reflects narrow, idealized operating conditions rather than real-world generalizability. Models trained on small datasets frequently fail when deployed in variable farm environments (10-30% accuracy degradation observed in cross-validation studies).

#### 5.1.2. Tier 2: Medium-Scale Datasets (500-5,000 images, n=42 studies)

- **Characteristics:** Single-farm deployments, moderate breed diversity (1-3 breeds), some environmental variation (indoor/outdoor, day/evening), collection over 1-6 months.

- **Typical ML Performance:** 85-95% (approaching performance ceiling due to feature engineering limitations)
- **Typical DL Performance:** 95-99% (e.g., [48] achieves 98.7% on 4,923 images using VGG16_BN with batch normalization)
- **Best-Suited Methods:** Lightweight deep learning architectures (VGG, MobileNet, EfficientNet) or ML ensembles.
- **Critical Transition Zone:** This range is where deep learning starts to outperform classical ML (typically with 1,000-1,500 images). Transfer learning from pre-trained ImageNet models is highly effective here.

#### 5.1.3. Tier 3: Large-Scale Datasets (5,000-50,000 images, n=24 studies)

- **Characteristics:** Multi-farm deployments, multi-breed populations (3+ breeds including indigenous varieties), diverse environmental conditions (all weather, seasonal variation, day/night cycles), collection over 6+ months.
- **Typical ML Performance:** 75-85% [33] (performance degradation due to increased intra-class variability and inter-class similarity across breeds).
- **Typical DL Performance:** 96-99% (e.g., [37] achieves 97% mAP on 27,849 images using Faster R-CNN for coat pattern identification)
- **Best-Suited Methods:** Deep architectures (ResNet-50/101/152, DenseNet-121/201, Inception-V3/V4) with data augmentation and regularization.
- **Performance Gap Widening:** Deep learning surpasses classical ML by 10-20 points as hierarchical features capture complex variations missed by manual engineering.

#### 5.1.4. Tier 4: Small-Scale Datasets (< 500 images, n=15 studies)

- **Characteristics:** Video surveillance data, temporal tracking requirements, extreme environmental variability (uncontrolled outdoor pastures, long-term monitoring >1 year), multi-camera systems
- **Typical ML Performance**: Not applicable (manual feature extraction infeasible for video-scale data)
- **Typical DL Performance:** 84-96% (e.g., [50] achieves 84% mAP on 505,800 video frames using YOLOv5 + ResNet101 for facial re-identification)
- **Best-Suited Methods:** Video-specific deep learning (YOLO for real-time detection, LRCN for temporal modeling, ResNet + ArcFace for re-identification, DeepSORT for tracking).
- **Unique Challenges:** Temporal consistency across frames, motion blur, extreme pose variations, long-term appearance changes (seasonal coat shedding, weight fluctuations), computational scalability for real-time processing.

To compare results across these dataset complexity tiers, we propose evaluating performance relative to tier-specific benchmarks:

- Tier 1 (Small-Scale) Benchmark: 95% threshold (simpler task, controlled conditions, lower generalization requirement)
- Tier 2 (Medium-Scale) Benchmark: 90% threshold (moderate complexity, some real-world variation)
- Tier 3 (Large-Scale) Benchmark: 85% threshold (high complexity, significant environmental and breed diversity)
- Tier 4 (Video-Scale) Benchmark: 80% threshold (extreme variability, temporal consistency requirements, real-time constraints)

Using this normalization framework, we can fairly compare studies across tiers:

1. Study A [121]: 100% accuracy on 217 muzzle images (Tier 1)
   - Interpretation: Exceeds Tier 1 benchmark by +5 percentage points. Strong performance, but limited generalizability due to controlled conditions.
2. Study B [48]: 98.7% accuracy on 4,923 muzzle images (Tier 2)
   - Interpretation: Exceeds Tier 2 benchmark by +8.7 percentage points. Excellent performance with moderate-scale real-world applicability.
3. Study C [50]: 84% mAP on 505,800 video frames (Tier 4)

- Interpretation: Exceeds Tier 4 benchmark by +4 percentage points. Strong performance given extreme video complexity and real-time processing requirements.

## 6. Ethical, Economic, and Legal Considerations

The integration of advanced machine learning (ML, which uses algorithms to identify patterns in data) and deep learning (DL, a subset of ML that relies on neural networks to model complex relationships) techniques into cattle identification systems presents both technical advances and important ethical, economic, and legal implications. Considering these aspects helps ensure technologies align with societal values, promote equitable access, and comply with regulations. This section explores these dimensions, with a focus on contexts such as Qurbani in Bangladesh, where livestock management meets religious and ethical norms [136].

### 6.1. Animal Welfare Implications

Non-invasive identification methods, such as image-based ML and DL techniques, represent a significant improvement over traditional physical tagging by minimizing stress and injury to cattle. Unlike ear notching or branding, which can cause pain, infection, and long-term discomfort, visual recognition systems—leveraging features like muzzle prints or coat patterns—require no physical contact and adhere to principles of humane treatment [137]. For instance, studies on muzzle matching with convolutional neural networks (CNNs, a deep learning architecture for image data) have shown high accuracy in cattle identification without handling, reducing animal stress compared to invasive methods [138]. These systems also enable continuous, remote monitoring using cameras or drones, supporting early detection of diseases or abnormal behaviors via behavioral analysis. For example, CNNs can process video feeds to identify signs of lameness or distress, allowing timely interventions that enhance herd health [139].

This shift also supports compliance with established animal welfare frameworks, such as the Five Freedoms (a welfare standard that provides animals with freedom from hunger, discomfort, pain, and fear and allows them to express natural behaviors). By reducing the need for frequent handling, DL-based systems promote free-range practices and lower cortisol levels in animals, as evidenced by studies showing reduced stress responses in monitored herds [140]. However, ethical deployment requires safeguards against over-surveillance, ensuring that monitoring enhances welfare without infringing on animals' natural behaviors.

### 6.2. Cultural Sensitivity

In regions like Bangladesh, cattle play a central role in cultural and religious practices such as Qurbani during Eid al-Adha. AI-driven identification must respect traditional values and community norms, as Qurbani emphasizes the animal's physical integrity and ethical treatment. Any visible alteration, such as branding or tagging, can diminish its sanctity [141]. Image-based techniques preserve the animal's appearance, enabling traceability without compromising aesthetic or spiritual standards. Non-intrusive facial or muzzle recognition verifies breed purity and health history during market transactions, fostering trust among farmers and buyers while honoring religious guidelines [142].

To further support cultural acceptance, implementation should use participatory design and gather feedback from religious leaders and local stakeholders. This approach builds trust in the technology. Transparent algorithms that explain decisions help avoid "black-box" perceptions in sacred practices. Integrating AI with traditional management, like combining digital records with oral herd histories, bridges generational knowledge gaps. This approach promotes cultural preservation and modernization [143].

### 6.3. Economic Considerations

Upfront costs for ML/DL systems, such as cameras and computational resources, may deter adoption. However, long-term savings in labor, disease prevention, and improved traceability deliver a strong ROI. Automated identification can reduce tagging errors by up to 30%. Large farms can cut operational costs by 15-20% due to streamlined herd management and fewer veterinary interventions [144]. Smallholder farmers—over 80% of Bangladesh's livestock sector—face accessibility challenges. Subsidized open-source models and mobile apps for image upload could lower barriers, while insurance premiums may drop 5-10% thanks to better traceability. These systems help prevent losses from theft or market misidentification, supporting rural livelihoods and strengthening supply chain resilience [145].

### 6.4. Legal and Regulatory Framework

The use of AI in livestock identification raises legal questions about data privacy, intellectual property, and compliance. Farm-level animal images and health records must be protected by new agricultural data rules, similar to the EU's GDPR, which requires consent and anonymization to prevent misuse, such as unauthorized sharing with insurers [146]. In Bangladesh, alignment with the Digital Security Act (2018) and new livestock traceability laws is essential to protect farmers from cyber risks [147].

Intellectual property rights for breed identification models, often built on proprietary datasets, require clear licensing. This enables open innovation and prevents agribusiness monopolization [148]. Compliance with international standards, such as ISO 22005 for food traceability (a globally recognized system that specifies requirements for tracking the flow of food and animal feed), requires certification for AI systems to ensure cross-border interoperability. Standardization bodies should develop ethical AI guidelines, including bias audits to prevent discriminatory outcomes, such as lower accuracy for indigenous breeds. Non-compliance can lead to fines or market exclusion. Proactive legal frameworks are needed to balance innovation with accountability.

### 6.5. Societal Impact

The societal effects of AI in cattle identification involve employment, equity, and community development. Automation may displace traditional roles, such as manual taggers, but it also creates jobs in tech maintenance, data analysis, and AI training. Rural areas could see 20-30% more skilled jobs through capacity-building programs [149]. However, digital divides persist, as about 40% of Bangladeshi rural households lack smartphones or reliable internet, risking increased inequalities.

To address these societal challenges, technology transfer initiatives, including government-led workshops and NGO partnerships, should provide inclusive training. Empower women and youth in livestock management. Support systems, like helplines and shared hardware cooperatives, help democratize access and foster social cohesion [150]. These technologies can help meet development goals and strengthen food security and rural economies. Still, they require oversight to ensure equitable benefits.

## 7. Conclusion

This The evolution of cattle identification techniques marks a turning point in livestock management. It enables improved biosecurity, optimized supply chains, and sustainable farming. Modern applications are shifting from traditional methods to advanced machine learning (ML) and deep learning (DL) approaches. These new methods offer superior precision, scalability, and efficiency. They deliver remarkable accuracy but require large datasets and more computational resources. Advancing identification technologies is essential for greater accuracy, operational efficiency, and animal welfare.

In Bangladesh, where livestock management is a valued tradition, cultural practices like Qurbani shape the industry. For those involved in the cattle market, advanced identification techniques such as facial recognition, gait analysis, and back pattern detection offer practical solutions. These tools boost transparency and traceability, enhancing trust and efficiency while respecting local customs. By adopting these non-invasive approaches, stakeholders can achieve business growth, uphold ethical standards, and reduce stress for animals.

Advanced machine learning (ML) and deep learning (DL) techniques have greatly enhanced the accuracy and efficiency of cattle identification. Methods like feature extraction, convolutional neural networks (CNN), and You Only Look Once (YOLO) object detection far surpass traditional approaches. While challenges remain—such as dataset size, environmental variation, and real-time processing—the deployment of these methods offers significant value. This impacts livestock management, biosecurity, and supply chain traceability.

Researchers and practitioners must prioritize building robust, representative databases and actively foster collaboration across animal sciences, computer science, and agricultural engineering. Balancing global standards with local needs, such as those in Bangladesh, is essential for developing humane and effective identification systems. Act now to ensure precision, scalability, and ethical practices at scale so cattle identification can drive sustainable, resilient, and responsible livestock production while respecting cultural and environmental values.

## 8. Future Work

Future Future work with machine learning (ML) and deep learning (DL) will aim to improve the detection of cattle breeds. This includes better ways to measure mixed-breed percentages using classification models with many possible categories. We will build and test image collections showing muzzle prints, coat patterns, and body shapes, and compare them with improved computer models like CNN-ResNet. This lets us better profile individual animals and breeds. We found that measuring body features helps non-invasively identify cattle and could help pick the best animals for fattening by linking what we see to how they grow. This work builds a solid base for future cattle identification research and innovation.

### 8.1. Development of Advanced DL Architectures

To address challenges in feature extraction under variable conditions, such as occlusions or lighting, future work should investigate transformer-based architectures. Vision Transformers (ViT) and related models excel at capturing long-range dependencies in images, offering improvements over traditional CNNs for tasks like muzzle or coat pattern recognition. Distillation-driven transformers, such as CattleDiT, and lightweight MobileViT hybrids could enable efficient, high-accuracy identification on resource-constrained devices, achieving up to 95% precision in noisy datasets. Ensemble methods combining CNNs with multi-head attention fusion, such as MHAFF, can further boost generalizability across breeds and environments.

### 8.2. Self-Supervised Learning for Data-Efficient Identification

Labeled cattle datasets are scarce. Self-supervised learning (SSL) paradigms provide a promising solution by pre-training models on unlabeled images or videos, reducing reliance on manual annotations. Techniques such as contrastive learning or masked autoencoders extract robust features from raw footage of herd movements or facial landmarks, which directly increases re-identification accuracy in multi-camera setups. Notably, recent applications of SSL for Holstein-Friesian re-identification and animal detection in constrained farms report 10-15% improvements in F1-scores. Integrating SSL with domain adaptation also mitigates biases toward specific breeds, helping ensure fair and accurate identification across global livestock populations.

### 8.3. Multi-Modal Data Fusion for Comprehensive Identification

Image-based methods are currently dominant. However, fusing multi-modal data—such as visual features with sensor-derived biometrics or genetic markers—can improve individual and breed identification reliability. Notably, decision-level fusion strategies that combine face, muzzle, and ear tag features have demonstrated superior performance in uncontrolled settings. By incorporating environmental metadata, such as weather impacts on coat visibility, future hybrid models could deliver a holistic profile. Key outcomes include enhanced identification accuracy, reliable disease risk assessment, and scalable, non-invasive traceability in supply chains.

### 8.4. Real-Time Identification via IoT and Edge Computing

Deploying identification systems in real-time farm environments will require integration with IoT ecosystems. Edge-enabled smart cameras enable on-device inference. Key outcomes include minimized latency in large herds, improved biosecurity through prompt anomaly detection, and enhanced feeding optimization using accurate breed or individual identification to inform nutritional allocation based on breed-specific needs.

### 8.5. Explainable AI for Trustworthy Identification

Explainable AI (XAI) techniques will drive adoption by farmers by clarifying identification decisions, building trust, and supporting ethical compliance—especially in cultural contexts like Qurbani. For instance, attention heatmaps in ViT models or SHAP-based feature attribution can visualize key regions such as unique muzzle ridges, directly addressing concerns about transparency. Importantly, XAI helps to debug biases and ensures equitable model performance across indigenous and commercial breeds, ultimately leading to increased user trust, fairer outcomes, and wider adoption.

### 8.6. Mobile and Accessible Platforms for Scalable Deployment

Developing mobile applications with on-device ML will democratize access for smallholder farmers, resulting in increased accuracy and speed of breed identification in remote areas. Utilizing tools such as TensorFlow Lite enables farmers to quickly capture and analyze images. By employing cloud-edge hybrids, the system can offload complex computations and seamlessly integrate with existing apps for easy herd logging. This approach will reduce the digital divide, and real-time feedback loops with crowdsourced data will continuously improve model quality, leading to better decision-making and farm productivity.

## CRediT Author Contribution Statement

Fayazunnesa Chowdhury: Conceptualization, Methodology, writing – original draft, writing – review & editing; Syed Md Galib: Conceptualization, Supervision, Writing – review & editing; Md Nasim Adnan: Investigation, Validation, Writing – review & editing; Md. Moradul Siddique: Methodology; Investigation; Data curation; Writing – original draft; Writing – review & editing; Md Robiul Karim: Investigation; Resources; Writing – review & editing; K M Tanvir Anjum: Writing – original draft; Writing – review & editing.

## Acknowledgement

This research was financially supported by the University Grants Commission (UGC) of Bangladesh.

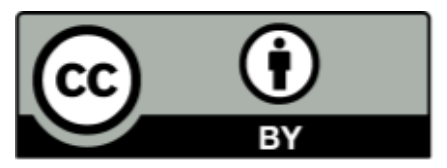